# SugarcaneNet: An Optimized Ensemble of LASSO-Regularized Pre-trained Models for Accurate Disease Classification

[1*]Md. Simul Hasan Talukder; Department of Electrical and Electronic Engineering, Dhaka University of Engineering & Technology, Gazipur, Bangladesh; Email: simul@duet.ac.bd

[2] Sharmin Akter; Department of Biomedical Engineering; Jashore University of Science and Technology, Jashore, Bangladesh; Email: sharmintalukder120@gmail.com

[3]Abdullah Hafez Nur; Department of Electronic and Telecommunication Engineering; International Islamic University Chittagong; Chittagong, Bangladesh; Email: engr.abdullahhafeznur@gmail.com

[4] Mohammad Aljaidi; Department of Computer Science; Faculty of Information Technology; Zarqa University, Zarqa, Jordan; E-mail: mjaidi@zu.edu.jo

[5] Rejwan Bin Sulaiman; School of Computer science and Technology; Northumbria University, UK, Email: Rejwan.sulaiman@northumbria.ac.uk

[6]Ali Fayez Alkoradees, Unit of Scientific Research, Applied College, Qassim University, Saudi Arabia; Email: alifk@qu.edu.sa

***Corresponding author**: Md. Simul Hasan Talukder; Email: simul@duet.ac.bd

**Abstract:** Sugarcane, a key crop for the world's sugar industry, is prone to several diseases that have a substantial negative influence on both its yield and quality. To effectively manage and implement preventative initiatives, diseases must be detected promptly and accurately. In this study, we present a unique model called SugarcaneNet that outperforms previous methods for automatically and quickly detecting sugarcane disease through leaf image processing. Our proposed model consolidates an optimized weighted average ensemble of seven customized and LASSO-regularized pre-trained models, particularly InceptionV3, InceptionResNetV2, DenseNet201, DenseNet169, Xception, and ResNet152V2. Initially, we added three more dense layers with 0.0001 LASSO regularization, three 30% dropout layers, and three batch normalizations with "renorm" enabled at the bottom of these pre-trained models to improve the performance. The accuracy of sugarcane leaf disease classification was greatly increased by this addition. Following this, several comparative studies between the average ensemble and individual models were carried out, indicating that the ensemble technique performed better. The average ensemble of all modified pre-trained models produced outstanding outcomes: 100%, 99%, and 99.45% for f1 score, precision, recall, and accuracy, respectively. Performance was further enhanced by the implementation of an optimized weighted average ensemble technique incorporated with grid search. This optimized "sugarcaneNet" model performed the best for detecting sugarcane diseases, having achieved accuracy, precision, recall, and F1 score of 99.67%, 100%, 100%, and 100%, respectively.

## 1. Introduction

Agriculture makes up around 6.4% of the world's economic activity. Out of the 226 countries in the globe, 9 of them have agriculture as the primary sector providing to their economy. China is the largest contributor, next to India, with the United States in third position. Sugarcane is a highly valuable crop that is recognized worldwide for its significant sucrose levels, which make up 80% of the world's sugar production, with an annual value of approximately US$150 billion [1]. Sugarcane, scientifically identified as Saccharum spp., is a type of perennial grass that belongs to the C4 category. It has the capacity to store huge quantities of sucrose in specialized cells called parenchyma storage cells [2]. It is an essential commercial crop that is employed for producing various by-products such as sugar, syrups, bagasse, fiberboard, and molasses, which are further used in the production of citric acid, ethyl alcohol and butyl alcohol [3]. The four main worldwide sugarcane producers are China, and Thailand, Brazil and India [4]. Sugarcane is a significant global crop that acts as the primary source of sugar and ethanol [5]. Sugarcane diseases in the sugar industry can lead to the eradication of crops, causing financial losses for small-scale farmers if not treated and noticed early. Sugarcane makes up 60% of the world's supply of raw sugar, while the remaining 40% originates from sugar beet [6].

But it is a matter of serious concern that the invasion of different diseases significantly hampered sugarcane production. Ratoon stunting disease (RSD) is a very economically impactful disease which damages sugarcane. It is caused by a bacterium named Leifsonia xyli subsp. xyli (Lxx), which resides in the xylem of the sugarcane plant [7],[8]. Sugarcane ratoon stunting disease (RSD) is widespread and causes harmful disease that affects sugarcane crops worldwide. It is caused by the bacterium Leifsonia xyli subsp. xyli (Lxx). Lxx is a pathogen that is required to cause disease in sugarcane, and previous study has documented certain physiological responses in sugarcane plagued by RSD [9]. Another Disease called Sugarcane mosaic disease (SMD) poses an important danger to the production of sugarcane [10]. At currently, the management of SMD mostly depends on the development of resistant cultivars through hybridization, a procedure that is time-consuming [11]. another disease named Sugarcane yellow leaf virus (SCYLV) is a highly widespread virus that causes disease in sugarcane on a global scale [12]. Sugarcane yellow leaf virus (SCYLV), a species of the Polerovirus genus in the Luteoviridae family, was initially identified in China in 2006 as the cause of sugarcane yellow leaf disease (YLD) [13]. The existence of SCYLV is widespread throughout most states in India where sugarcane production takes place. A virion consists of 180 coat protein units and has a diameter of 24-29 nm. The SCYLV genome is monopartite and consists of a single-stranded (ss) positive-sense (+) linear RNA molecule that is roughly six kb in size [14],[15]. Agriculture, as one of the most ancient industries, acts as the foundation of various economies [16]. According to a UN report, as the world's population is expected to exceed 9.8 billion by 2050, the need for efficient and environmentally friendly agricultural methods is more important than ever [17]. The all over the world landscape of agriculture is always changing due to challenges and improvements that impact the efficiency and durability of farming methods [18]. Technology plays a vital part in driving improvements in agriculture, influencing areas such as crop cultivation and disease management [19],[20].

The study analyses the vital connection between agricultural production and advancements in technology, with an emphasis on sugarcane agriculture, a crucial cash crop across different Indian regions. Agrarian nations rely heavily on agriculture, illustrating the important impact that any changes in agricultural productivity may have on their financial situation and daily activities [21]. Many agrarian countries have an important share of their workforce engaged in agriculture-related activities. A decrease in agricultural productivity has a substantial impact on daily life. Production losses are greatly affected by bad weather conditions, insufficient irrigation, unsuitable crop choices, plant diseases, and a lack of modern farming infrastructure [22]. To tackle these issues, we need innovative approaches that can boost crop production, better disease identification, and optimize resource utilization.

Deep learning and Internet of Things (IoT) technologies have the ability to substantially transform agricultural operations. IoT and deep learning integration may successfully improve crop yield and product quality [23]. These technologies assist in developing intelligent systems that can make informed decisions, increasing the efficiency and efficacy of farming operations. Technological advancements in agriculture, such as context-aware irrigation systems, precision farm gear with GPS and GIS, and hyperspectral imaging for crop analysis, have grown more popular [24]. The use of UAVs with deep learning for yield estimation and disease detection illustrates the potential of integrating advanced technologies into agricultural operations. Images and data collected by UAVs from the ground can significantly assist in accurately forecasting crop production. The study validates the assertion that advances in technology will lead to improved sugarcane yields [25].

Deep learning has revolutionized picture identification, image classification, and other areas that necessitate the handling of huge amounts of data. The use of deep learning in plant disease detection has revolutionized the methods specialists use to assess and make judgments [26]. CNN is a widely used technique for presenting intricate concepts, utilizing a substantial amount of data for pattern recognition tasks. A Convolutional Neural Network (CNN) is a specialized technique used for image recognition or classification, containing pooling layers, fully connected layers and multiple convolution layers [27]. Research on sugarcane plant diseases concentrates on the biological features of the diseases.

Implementing deep learning technologies in agriculture is hindered by the absence of suitable datasets for training models. Deep learning technologies, such as Convolutional Neural Networks (CNNs) and Deep Belief Networks (DBNs), have been suggested for detecting plant diseases and infestations [28]. These techniques have demonstrated favorable results in identifying and detecting lesions from digital images [29],[30]. Transfer learning enables pre-trained models to be tailored for new tasks with less data, but generating specialized datasets for agricultural applications is still an exhausting and time-consuming endeavor. This is particularly reliable for identifying diseases in crops such as sugarcane, as the differences in symptoms of disease and environmental factors needed for the development of strong, immediate diagnostic tools.
The present study provides a unique weighted average ensemble approach of customized, and LASSO regularized transfer learning models called SugarcaneNet to identify five types of sugarcane diseases using leaf image processing. In the study, we modified the seven pre-trained models, namely InceptionResNetV2, ResNet152V2, EfficientNetB0, InceptionV3, DenseNet201, Xception, and DenseNet169, adding LASSO, dropout, and renormalization regularization to avoid overfitting, and those were fitted using early stop optimization techniques to abstain from overfitting in the next level. Later, prominent 22-average ensemble models were designed, implemented, and evaluated to enhance performance. Finally, a weighted average ensemble of the whole seven models was developed that outperformed the others. Moreover, the weight values were tuned by grid search techniques.

The paper's major contributions are:

- Customizing seven pre-trained models.
- Addition of three densely with 0.0001 LASSO regularization.
- Incorporation of three renormalizations along with batch normalization.
- Introduction of 30% dropout layer.
- Comparative study of the seven pre-trained models.
- Development of 22 average ensemble models and analysis of their performances.
- Implementation of weighted average ensemble model.
- Optimization of weights with grid search techniques.
- Proposing proficient “SugarcaneNet” model for sugarcane leaf diseases classification.

Section 2 presents Literature survey on sugarcane diseases classification. The materials and methods are explained in Section 3. Results and discussion are illustrated in Section 4. The conclusion and future work are presented in Section 5.

## 2. Literature Survey

Artificial intelligence has emerged as an emerging technology in the agricultural field. Different research is carried out in disease classification [46], pest identification [47-48], nutrient deficiency detection [49], and so on. Similarly, several studies have also accomplished sugarcane disease classification. Upadhye et al. [31] proposed an adapted deep learning Convolutional Neural Network (CNN) approach to identify sugarcane diseases, which handles the important problem of disease detection in sugarcane fields that can lead to economic losses for farmers. They devised an approach that categorized sugarcane images as healthy or unhealthy with an impressive accuracy of 98.69%. The research utilized a well-rounded dataset that covered various ago-climatic conditions, lighting variations, and plant densities across India to ensure thorough detection of diseases. The CNN approach utilized convolutional, ReLU, pooling, and fully connected layers tailored for the classification of four sugarcane diseases and a healthy plant category. The research demonstrated great accuracy in classifying diseases and helped create a user-friendly web-based application for farmers to quickly identify diseases, demonstrating the promise of machine learning in agricultural disease control.

Kotekan et al. [32] developed an automated system to detect sugarcane leaf diseases by using a ConvNet (CNN) deep learning technique. Through the application of a dataset including 13,842 visuals categorized into 7 categories, which included healthy and diseased leaves, they effectively achieved high accuracy in identifying diseases. They acquired image datasets, standardized them to a 96x96 resolution, and employed convolution layers for extracting features. The classification process utilized completely connected layers. Initial testing showed a performance decline to 31.4% under different conditions. Nevertheless, after creating a better dataset with 2,940 visuals across six classes, the model's efficacy was bolstered. The study demonstrated that Convolutional Neural Networks (CNNs) are outstanding in identifying sugarcane diseases and have the potential to be utilized in real-world agriculture. It emphasized the importance of high-quality and diverse datasets for the model to perform effectively. Vignesh and Chokkalingam [33] proposed an EnC-SVMWEL method for detecting and classifying sugarcane leaf diseases based on leaf images. The research uses a dataset of 2940 sugarcane leaf pictures, consisting of 928 healthy and 2012 diseased leaves categorized into 5 disease classes. It employs the DenseNet201 architecture for feature extraction and a new SVMWEL classifier for disease classification. The EnC-SVMWEL model demonstrated outstanding efficacy with a classification accuracy of 97.45%, surpassing other methods in criteria including precision, recall, and f-measure. The analysis was carried out using MATLAB on a Windows Intel i3 system with 6GB RAM. The dataset was divided into 80% for training and 20% for testing. Garg et al. [34] created a convolutional neural network (CNN) which includes Long-short term memory (LSTM) to detect Sugarcane Brown Spot (SBS) disease early and categorize its severity. The model obtained a binary detection accuracy of 98.11% in distinguishing between healthy and diseased sugarcane leaves using a dataset of 20,000 images from Punjab sugarcane fields. It also split the severity of SBS into four stages with a multi-classification accuracy of 93.87%. This method exceeded modern methods in detecting SBS and evaluating its severity, which has significant implications for agriculture as it enables quick disease control and distribution of resources.

Banerjee et al. [35] developed a hybrid CNN-SVM model to forecast Grassy Shoot Disease (GSD) severity in sugarcane crops with high accuracy. With a dataset of 2925 sugarcane plant pictures, the investigators conducted preprocessing, feature extraction using a three-layer CNN, and classification with an SVM integrating regularization. The model obtained an 81.53% overall accuracy, with precision, recall, and F1-score values showing variability throughout severity levels, with the maximum precision of 85.37% seen at Severity Level 8. The method outperforms conventional visual assessment approaches, demonstrating its capability to improve decision-making in sugarcane production for GSD management. Dhawan et al. [36] developed a CNN-LSTM ensemble model to evaluate severity levels of sugarcane downy mildew disease utilizing a dataset of annotated sugarcane leaf images with severity levels ranging from 1 to 5. The model incorporates CNN's spatial feature extraction with LSTM's temporal analysis to forecast disease severity. Assessed on a dataset divided into 7000 training and 3000 testing pictures, the model obtained an overall accuracy of 94.16%, as well as high precision, recall, and F1 scores. The technique demonstrates promise for automating the detection of sugarcane downy mildew, indicating an important advance in agricultural data analysis and disease control methods.

Sharma and Kukreja [37] created a deep learning model utilizing a multi-layer perceptron (MP) to identify Sugarcane red rot (SRR) disease. The model was developed utilizing a dataset of 2000 sugarcane images from the Punjab region of India. The goal of the model is to assess the health status of a plant and categorize the severity of SRR disease into five levels. The study showed a binary classification accuracy of 99.12% for distinguishing between healthy and SRR diseased individuals. It also obtained an average accuracy of 98.94% for multi-classification across different SRR severity levels, with the maximum accuracy of 99.15% for the SI level of the disease. The technique exceeds previous attempts in detecting sugarcane diseases, representing a notable progression in the field.

Aruna et al. [38] developed the Inception Nadam L2 Regularized Gradient Descent (NLRGD) CNN model to categorize sugarcane diseases. They utilized the Sugarcane Disease Dataset from KAGGLE, which has 3000 leaflets categorized into Bacterial Blight, Red Dot, and healthy plants. The dataset was split into 2700 training pictures, 150 validation images, and 150 testing images. The NLRGD model, which includes 3 x 3 convolutional layers, Inception networks, L2 regularization, and NADAM optimization, was compared to usual CNN designs such as ResNet, GGG19, DenseNet, and Xception. The NLRGD model demonstrated outstanding performance with 96.75% accuracy, 96.62% precision, 96.25% recall, and a 96.76% f1 Score, confirming that it is useful in identifying and classifying sugarcane diseases.

Tanwar et al. [39] performed a study at the Sugarcane Research Institute in Shahjahanpur, Uttar Pradesh, India, to look into the incidence of sugarcane grassy shoot disease. The research showed infection rates ranging from 6% to 28% across various sugarcane varieties. The research employed convolutional neural networks (CNN) to forecast and categorize this condition, with a 96% accuracy level. The investigation utilized a dataset of 1000 photos of red-rot-infected sugarcane leaves collected from internet repositories such as GitHub and Kaggle. It implemented a five-step CNN model involving dataset collection, distribution, image pre-processing, feature extraction, and analysis. The model was evaluated using a 20:80 split between test and training sets. It consisted of 13 layers and utilized max-pooling and convolutional methods for feature extraction. The CNN model exhibited superior performance compared to SVM and KNN models, with a 96% accuracy in illness prediction. It also achieved F1 scores of 96.5% for healthy leaves and 96% for unhealthy leaves, outperforming the other models with an average accuracy of 92% on test pictures. The research illustrates the effectiveness of CNN in diagnosing sugarcane grassy shoot disease, despite constraints such as dataset size and disease specificity.

Tanwar et al. [40] created a hybrid model which combines Convolutional Neural Networks (CNN) and Support Vector Machines (SVM) to forecast the severity of leaf smut in sugarcane. The disease severity was divided into four levels: mild, average, severe, and profound utilizing a dataset of 950 photos from Mendeley and Kaggle. The model exhibited high accuracy rates of 98% for mild and severe categories, and 97% for average and profound categories. The performance indicators, particularly recall, precision, and F1-score, revealed high values for each category, confirming the model's effectiveness in categorizing the severity of sugarcane leaf smut. The method exhibited great accuracy compared to other models and emphasized the possibility of combining CNN and SVM in diagnosing agricultural diseases and predicting their severity.

Maurya et al. [41] developed an improved VGG16 model designed specifically for identifying sugarcane leaf diseases. They utilized a distinct dataset classified into "Healthy", "Bacterial Blight", "Red-Rot", and "Rust". They expanded the original dataset of 400 photos to 6400 utilizing different deep learning approaches, resulting in an important classification accuracy of 94.47% on an NVIDIA DGX server. The technique surpasses prior models such as Inception V3 and VGG19, confirming its capability to assist farmers in early disease diagnosis and management.

Hernandez et al. [42] created deep-learning models to identify sugarcane diseases based on leaf images. Their study involved 16,800 training pictures, 4,800 validation pictures, and 2,400 testing pictures. The InceptionV4 algorithm has the highest accuracy of 99.61%, exceeding VGG16, ResNetV2-152, and AlexNet, which had accuracies of 98.88%, 99.23%, and 99.24% respectively. The optimization techniques used greatly improved the model's performance, shown by the F1 scores. The study, despite its limited dataset, confirms the effectiveness of deep learning in identifying sugarcane diseases. It highlights the necessity for further research using larger datasets and real-world experimentation.

Tomar and Chaurasia [43] performed a literature study on the use of computational algorithms to identify diseases in sugarcane plants. They investigated different image processing methods and deep learning strategies by analyzing over 1400 articles from different sources. The study concluded that developing an automated system to identify sugarcane diseases is essential for improving agricultural productivity and lowering disease control costs.

Earlier, Cuimin Sun et. al. [44] proposed the SE-ViT hybrid model to diagnose sugarcane leaf disease. The study included five types of diseases, such as healthy, red stripe, ring spot, brawn stripe, and bacteria, that were collected from the plant village dataset and further improved. The research achieved 89.57% accuracy.

Recently, Daphal et. al. [45] prepared a novel real-time dataset having five types of sugarcane leaves, such as red, mosaic, yellow, healthy, and rust, and conducted a comparative study to classify the dataset with transfer learning and an ensemble approach. The ensemble of a sequential CNN model and another deep CNN model with spatial attention achieved 86.53% accuracy. The dataset is very recent, and it included five real-field Sugarcane datasets, including mosaics that were absent from other datasets.

In the literature survey, most research has focused on improving the performnace of already-existing models. A notable finding of these studies is that a large percentage of them center on the binary categorization of sugarcane diseases, that is, the differentiation of healthy plants from unhealthy ones. Certain sugarcane diseases, including downy mildew, brown spot, or red rot, have been the only focus of some researchers, while three different diseases—Bacterial Blight, Red Dot, and healthy plants—have been included in other studies. Additionally, some studies have increased the number of groups in the classification to four: Rust, Red-Rot, Bacterial Blight, and Healthy. Further research has focused on the severity level classification of sugarcane illnesses, frequently classifying them into five groups according to their outward appearances, such as healthy, red stripe, ring spot, brown, and so on.

Several approaches, including deep learning, transfer learning, and convolutional neural networks (CNN), as well as classic and hybrid approaches, have been used in these research projects. Studies that combined CNN with spatial domain CNN have even investigated ensembles. The use of more sophisticated strategies, such as modified regularized transfer learning, and ensemble approaches, such weighted ensemble, all possible average ensemble, and grid search-based tuned weighted average ensemble, has not yet been widely adopted.

Despite the extensive study, the accuracy attained is still quite low; the most recent work [45] only managed to reach 86.53%. To address these issues and improve performance even more, we suggest the "sugarcaneNet" model, which was created using a comparable dataset and has shown excellent performance.

## 3. Methodology

The entire process of this study has been illustrated in figure 4 for ease of comprehension. This method involves a variety of approaches and is made up of multiple phases that are worked out once at a time.

### 3.1. Dataset

The sugarcane leaf disease dataset has been collected from a trustworthy Mendeley repository [50]. There are five main categories in the manually gathered sugarcane leaf disease image dataset: rust, yellow disease, redrop, mosaic, and healthy. To ensure representativeness and diversity, a variety of smartphone combinations were used to take these pictures. There are 2569 images across all categories in the collection. The collection activities were carried out in Maharashtra, India, which reflects the sugarcane production setting specific to the region. The extremal appearance of the sugarcane leaves is shown in Figure 1, and the details of the distribution of the classes are summarized in Table 1.

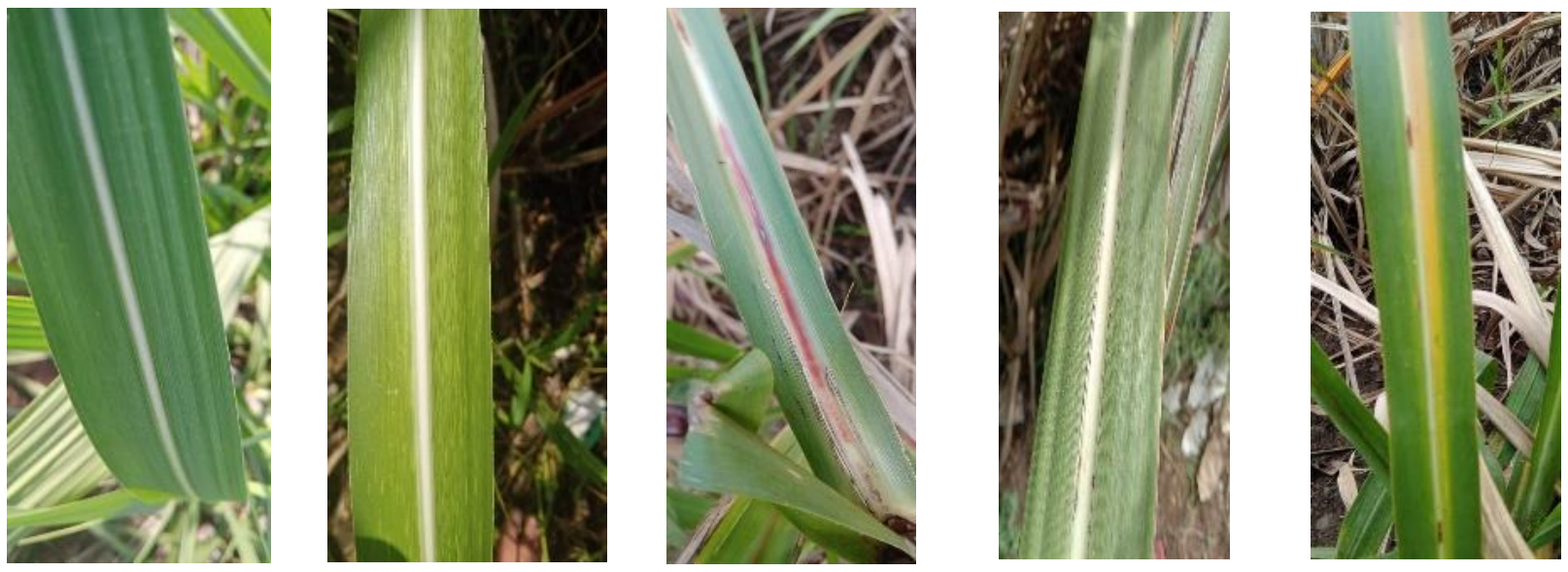

Figure 1. Different views of sugarcane leaves.

Table 1. Data distribution.

| Class Name | Number of Images |
|---|---|
| Healthy | 520 |
| Mosaic | 514 |
| RedRot | 519 |
| Rust | 505 |
| Yellow | 511 |
| Total | 2569 |

### 3.2. Customized TL Models

The concept of transfer learning refers to the process of adapting or repurposing a model that has been trained in one job to another that is similar, as shown in Figure 2 [51]. This is especially helpful in situations where there is a shortage of labelled data for the new task because the pre-trained model has already picked up relevant features from the "ImageNet" dataset, having 14 million images [52]. The TL technique enhances the learning of models, providing logical, faster, and better performance [53]. Its work can be presented mathematically for better understanding.

Assume, Task P and Task Q are our two assignments. A pre-trained model that was trained on Task P is also available.

The symbol $f_P(\,)$ represents the pre-trained model. The prediction is $Y_P$, where X is the input of Task P. Thus, it can be expressed as-

$$Y_P = f_P(X) \qquad (1)$$

The input for Task Q is $X_Q$, and the true labels are $Y_Q$. Due to insufficient data, models that are trained directly tend to overfit.

For this reason, to accomplish the second task, which is called the "adapted model denoted by $f_{PQ}$," we updated the pre-trained model using the prior information. This model's predction is $Y_{PQ}$.

$$Y_{PQ} = f_{PQ}(X_Q) \qquad (2)$$

By using the Q task label to fine-tune the pre-trained model's parameters, the adaptation process is carried out. A loss function measuring the degree to which the ground truth labels for Task B deviate from the updated model's predictions can be reduced to achieve this. The optimization can be formulated by

$$min_{\theta_{PQ}} L(Y_Q, f_{PQ}(X_Q; \theta_{PQ})) \quad (3)$$

Where L is the loss function and $\theta_{PQ}$ are the parameters of adapted model.

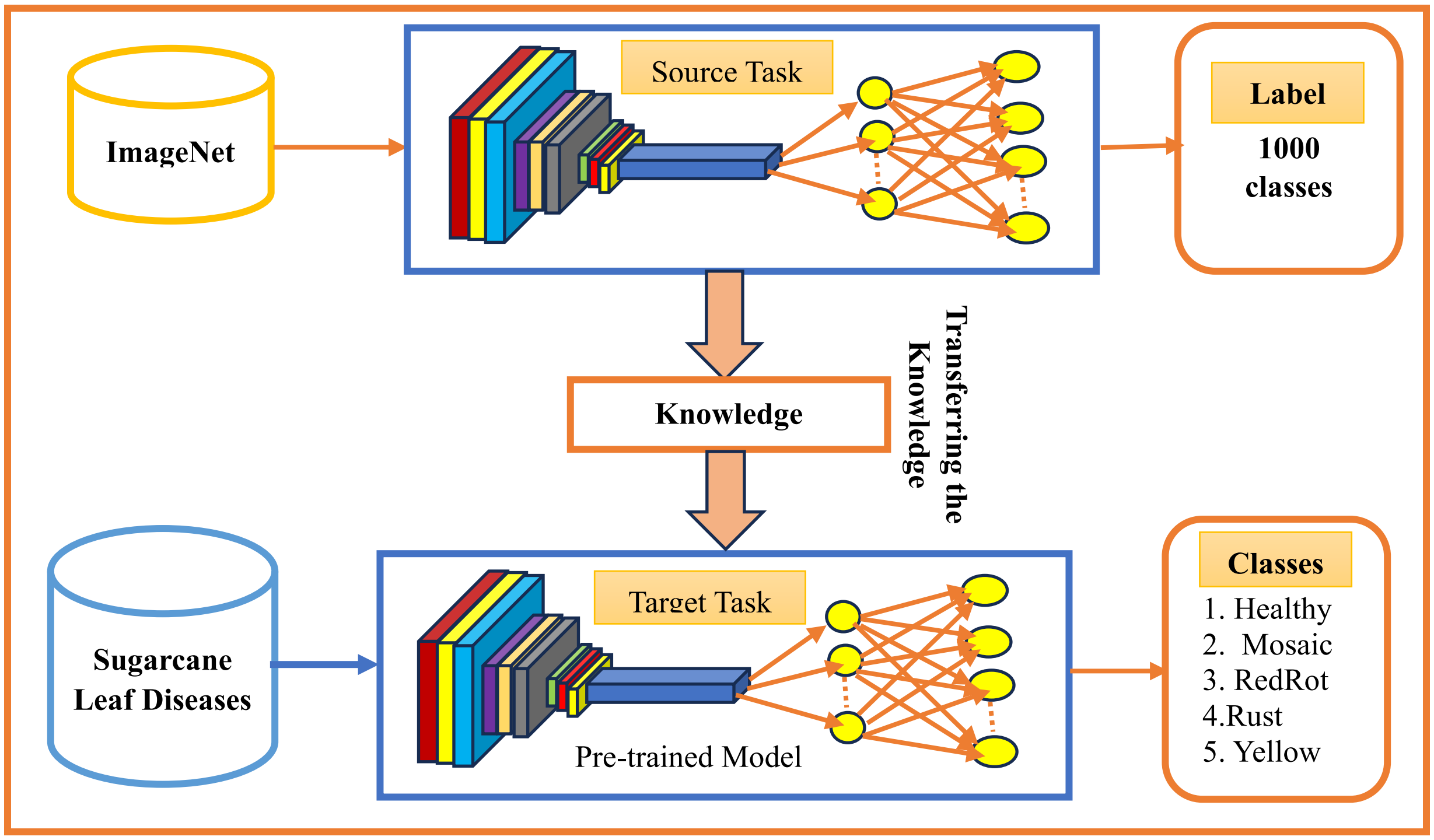

Figure 2. The concept of transfer learning.

In this work, we modified seven pre-trained models: InceptionV3, InceptionResNetV2, DenseNet201, Xception, DenseNet169, EfficientNetB0, and ResNet152V2, substituting the classification layers with four dense layers, three dropout layers, and three BatchNormalization Layer shown as the second portion "Customized Pre-trained Models" of the figure 4. The first three dense layers were improved through the application of Least Absolute Shrinkage and Selection Operator (LASSO) regularization, using a regularization parameter of 0.0001. These layers also used the Relu activation function to add non-linearity and make feature extraction easier. The number of trainable and non-trainable parameters are listed in Table 2.

### 3.2.1. Dense Layer

Dense Layer-also referred to fully connected layer is a basic block of neural network. It creates a densely connected network by joining all the neurons in the present layer to all the neurons in the previous layer. Mathematically, the output of the dense layer can be determined as follows if $X$ stands for the input to the dense layer, $W$ for the weights, and $b$ for the bias:

$$output = Activation\ (W * X + b) \quad (4)$$

In our work, SoftMax was utilized in the final dense layer, while ReLU activation was applied in the first three. The first three dense layers were incorporated with Least Absolute Shrinkage and Selection Operator (LASSO) regularization.

### 3.2.2. LASSO (L1 regularization)

L1 regularization in a dense layer indicates that the loss function is trained with an extra penalty based on the absolute value of the weights. Because it forces many of the weights to zero, this penalty promotes sparsity in the weights. This can lessen overfitting and aid in feature selection. Due the L1 regularization, the loss function of dense layer becomes as follows as-

$$Loss = Orginal_loss + \lambda \sum_{i,j} |w_{i,j}| \quad (5)$$

The original loss function, such as categorical cross-entropy for classification problems, is represented by the variable original loss. The regularization parameter, λ, regulates the degree of regularization. The weight that connects neuron $i$ in the current layer to neuron $j$ in the following layer is represented by the symbols, $w_{i,j}$.

### 3.2.3. Drop Out Layers

When dropout is used, a subset of neurons is chosen at random, and each training iteration, their outputs are set to zero. This enables the network to learn more resilient features and keeps the model from becoming overly dependent on any one group of neurons. We used three dense layers with the parameter of 0.3 that's means 30% of neurons became null after the preceding dense layer.

### 3.2.4. Batch Normalization

Batch normalization is a method that normalizes each layer's activations in a mini batch to enhance the training of deep neural networks. Batch normalization normalizes the activations of every layer for every mini batch during training by deducing the mean and dividing by the standard deviation. This is carried out separately for every feature dimension. The normalized activations are then scaled by the learnt parameter beta and shifted by the learned parameter gamma after normalization. These parameters are acquired during training, which enables the model to modify the normalization in an adaptive manner.

In our study, we also added renormalization step which is used to rescale the normalized activations to a target mean and standard deviation. The entire process can be presented by the following mathematical operation.

Assume, the standard deviation and meaning of each feature dimension for the entire mini batch are $\sigma$, μ respectively.

$$\mu = \frac{1}{m} \sum_{i=1}^{m} x_i \quad (6)$$

$$\sigma^2 = \frac{1}{m} \sum_{i=1}^{m} (x_i - \mu)^2 \quad (7)$$

$$\hat{x} = \frac{x-\mu}{\sqrt{\sigma^2-\epsilon}} \quad (8)$$

$$\hat{r} = \frac{d_{max}}{||\sigma||_2}, \frac{r_{max}}{||\mu||_2} \quad (9)$$

With the learning parameters and $\alpha$, scale and shift the normalized activations $\beta$

$$y = \gamma . r . \hat{x} + \beta \quad (10)$$

Where, $\hat{x}$ is the normalized input.
$\epsilon$ is a small constant added to the variance to avoid division by zero.
$\gamma$ and $\beta$ are learned parameters.
$r_{max}$ and $d_{max}$ are hyperparameters controlling the maximum allowed values for the L1 norms of $\mu$ and $\sigma$, respectively.
Finally, the output $y$ is passed to the next layer in the neural network.

Table 2. Trainable and non-trainable parameters of the modified pre-trained models.

| Model Name | Trainable Parameters | Non-Trainable |
|---|---|---|
| EfficientNetB0 | 4182465 | 43136 |
| InceptionResNetV2 | 54483877 | 61664 |
| DenseNet169 | 12708549 | 159520 |
| InceptionV3 | 22041573 | 35552 |
| Xception | 21080173 | 55648 |
| DenseNet201 | 18349765 | 230176 |
| ResNet152V2 | 58461125 | 144864 |

### 3.3. Ensemble Learning

Deep neural networks are extremely versatile and able to approximate any mapping function given training data and understand complicated correlations between variables since they are nonlinear models that learn through a stochastic training mechanism. Because of their sensitivity to the details of the training data and random initialization, this flexibility has a downside. Every training session, they might generate a distinct set of weights. With tiny datasets, this is especially true. Different predictions are made by the models with these various weights. Stated differently, there are a lot of variances in neural networks. Ensemble learning has proven to be an effective strategy for resolving the high variation issue [54].

We have created 21 average ensemble models in our study, which we achieved by combining two models from a set of seven customized pre-trained models with one all-model's average ensemble model, for a total of 22 ensemble learning approaches.
Since there are a total of 7 customized pre-trained models, the number of combinations of choosing 2 from them can be calculated as follows-

$$7_{C_2} = \frac{7!}{(7-2)!2!} = 21$$

And 1 is an all-model average ensemble. The total is 22. The final proposed "SugacaneNet2024" is designed with tuned weighted average ensemble learning approaches.

### 3.3.1. Average ensemble learning

The averaging ensemble method is a widely used technique in ensemble learning that involves averaging the predictions from different models. The tactics of average ensemble learning that is used in our approach are shortly summarized in algorithm 1. The predictions of each model were summed up and were chosen as the labels of the highest argument.

Algorithm 1. Average Ensemble Method

```
Input:   models= [model1, model2, -------model-n]
         X_test = test data
Output: Ensemble prediction
preds_list = []
For i=1 to n do
    For j=0 to n do
        preds = model.predict(X_test)
        APPEND preds TO preds_list
    End
    preds_array = CONVERT_TO_NP_ARRAY (preds_list)
    summed = SUM (preds_array) ALONG AXIS=0
    ensemble_prediction = ARGMAX (summed) ALONG AXIS=1
End
```

### 3.3.2. Tunned Weighted Average Ensemble Learning Model

A version of the ensemble learning technique known as a Tuned Weighted Average Ensemble Learning Model combines predictions from several base models using weighted averages; each model's weights are optimized or tuned [49-55]. The pseudocode of weighted average ensemble learning is presented in Algorithm 2. In our study, we chose a grid search space for each model from 0.0 to 0.4. The increment in weight was 0.1. The clear picture is shown in Table 3.

The total number of permutations of those weight parameters can be calculated as follows as-

$$7x7x7x7x7x7x7=78125$$

The outcome of the grid search is shown in Figure 3. The permutated weighted vector of [0.0, 0.1, 0.4, 0.0, 0.1, 0.4] yields the highest performance for which SugarcaneNet has been designed.

Table 3. Space of weights values.

| W1 | W2 | W3 | W4 | W5 | W6 | W7 |
|---|---|---|---|---|---|---|
| 0.0 | 0.0 | 0.0 | 0.0 | 0.0 | 0.0 | 0.0 |
| 0.1 | 0.1 | 0.1 | 0.1 | 0.1 | 0.1 | 0.1 |
| 0.2 | 0.2 | 0.2 | 0.2 | 0.2 | 0.2 | 0.2 |
| 0.3 | 0.3 | 0.3 | 0.3 | 0.3 | 0.3 | 0.3 |
| 0.4 | 0.4 | 0.4 | 0.4 | 0.4 | 0.4 | 0.4 |

| | wt1 | wt2 | wt3 | wt4 | wt5 | wt6 | wt7 | acc |
|---|---|---|---|---|---|---|---|---|
| **5500** | 0.0 | 0.1 | 0.3 | 0.4 | 0.0 | 0.0 | 0.0 | 98.348018 |
| **5501** | 0.0 | 0.1 | 0.3 | 0.4 | 0.0 | 0.0 | 0.1 | 98.788546 |
| **5502** | 0.0 | 0.1 | 0.3 | 0.4 | 0.0 | 0.0 | 0.2 | 99.008811 |
| **5503** | 0.0 | 0.1 | 0.3 | 0.4 | 0.0 | 0.0 | 0.3 | 99.118943 |
| **5504** | 0.0 | 0.1 | 0.3 | 0.4 | 0.0 | 0.0 | 0.4 | 99.118943 |
| **...** | ... | ... | ... | ... | ... | ... | ... | ... |
| **5652** | 0.0 | 0.1 | 0.4 | 0.0 | 0.1 | 0.0 | 0.2 | 99.229075 |
| **5653** | 0.0 | 0.1 | 0.4 | 0.0 | 0.1 | 0.0 | 0.3 | 99.559471 |
| **5654** | 0.0 | 0.1 | 0.4 | 0.0 | 0.1 | 0.0 | 0.4 | 99.669604 |
| **5655** | 0.0 | 0.1 | 0.4 | 0.0 | 0.1 | 0.1 | 0.0 | 99.008811 |
| **5656** | 0.0 | 0.1 | 0.4 | 0.0 | 0.1 | 0.1 | 0.1 | 99.229075 |

Figure 3. Snapshot of grid search outcome and tuned weighted values.

Algorithm 2. Weighted Average Ensemble Method

1 Input: models= $[model_1, model_2 \dots \dots \dots model_n]$
X_test = test data
W= $[W_1, W_2 \dots \dots \dots .. W_n]$
2 Output: Ensemble prediction
3 preds_list = []
4 **For** i=1 **to** n **do**
5 **For** j=0 **to** n **do**
6 preds = model.predict(X_test)
preds=preds*W
7 APPEND preds TO preds_list
8 **End**
9 preds_array = CONVERT_TO_NP_ARRAY (preds_list)
10 summed = SUM (preds_array) ALONG AXIS=0
11 ensemble_prediction = ARGMAX (summed) ALONG AXIS=1
12 **End**

### 3.5. Proposed SugarcaneNet

Our proposed SugarcaneNet model, which makes use of a tuned weighted average ensemble of seven customized and regularized pretrained models, is an example of an expertly constructed ensemble learning technique. Specifically developed for the purpose of using leaf image processing to classify sugarcane diseases, this ensemble model provides robustness and accuracy and improved disease detection. The flow diagram of the model and its architecture are pictorially presented in Figures 4 and 5 respectively.

The first step in the preprocessing stage was a comprehensive statistical examination of the dataset using OpenCV's image processing capabilities as well as the panda’s libraries. To maintain uniformity throughout the collection, all images were uniformly downsized to 224x224 dimensions. The dataset was then split into training and testing sets at a 70:30 ratio, providing the foundation for the creation of the model that followed.

As we moved into the model creation phase, we started with seven pre-trained models: DenseNet201, Xception, InceptionV3, EfficientNetB0, InceptionResNetV2, and ResNet152V2. Extra layers were added to these models: three batch normalization layers with renom=True, three dropout layers with a 30% probability, and three dense layers using 0.0001 Least Absolute Shrinkage and Selection Operator (LASSO) regularization. By penalizing features and parameters within the layers, these improvements successfully reduced overfitting.

Using distinct training and test datasets, the customized models were put through rigorous training, testing, and assessment procedures. To optimize the training and validation performance, Early Stopping techniques with monitor='val_loss', patience=7, and restore_best_weights=True were employed that aided in reducing overfitting. To ensure accuracy and robustness, performance was evaluated using a range of assessment indicators. We then created 22 average ensemble models, the first 21 of which were made up of pairings between two of the original seven models. The average of all the individual models was the last average ensemble model. Finally, we developed a weighted average ensemble strategy to improve the performance of the model even further. For each model, we found the ideal weights by applying grid search techniques. As shown in Figure 3, these optimized weights were subsequently used in the SugarcaneNet model's design.

We used precision, recall, F1 score, ROC curve, accuracy curve, loss curve, and confusion matrix studies for a thorough assessment. After carefully examining the data, it was clear that SugarcaneNet performed better than the other models in terms of disease classification.

### 3.6. Model Evaluation

Model evaluation is an important step for machine learning modelling. To find out how well a trained model generalizes new data, one must evaluate the model's performance. In our study, we have applied the following metrics in table 4 to assess SugarcaneNet.

**Accuracy**: The percentage of cases out of all instances that are correctly classified.

**Precision:** The percentage of actual positive cases among all positive cases that have been classified.

**Recall**: The percentage of real positive examples among all real positive examples that were appropriately classified.

**F1 Score**: The harmonic means of recall and precision, which offers a fair assessment of each.

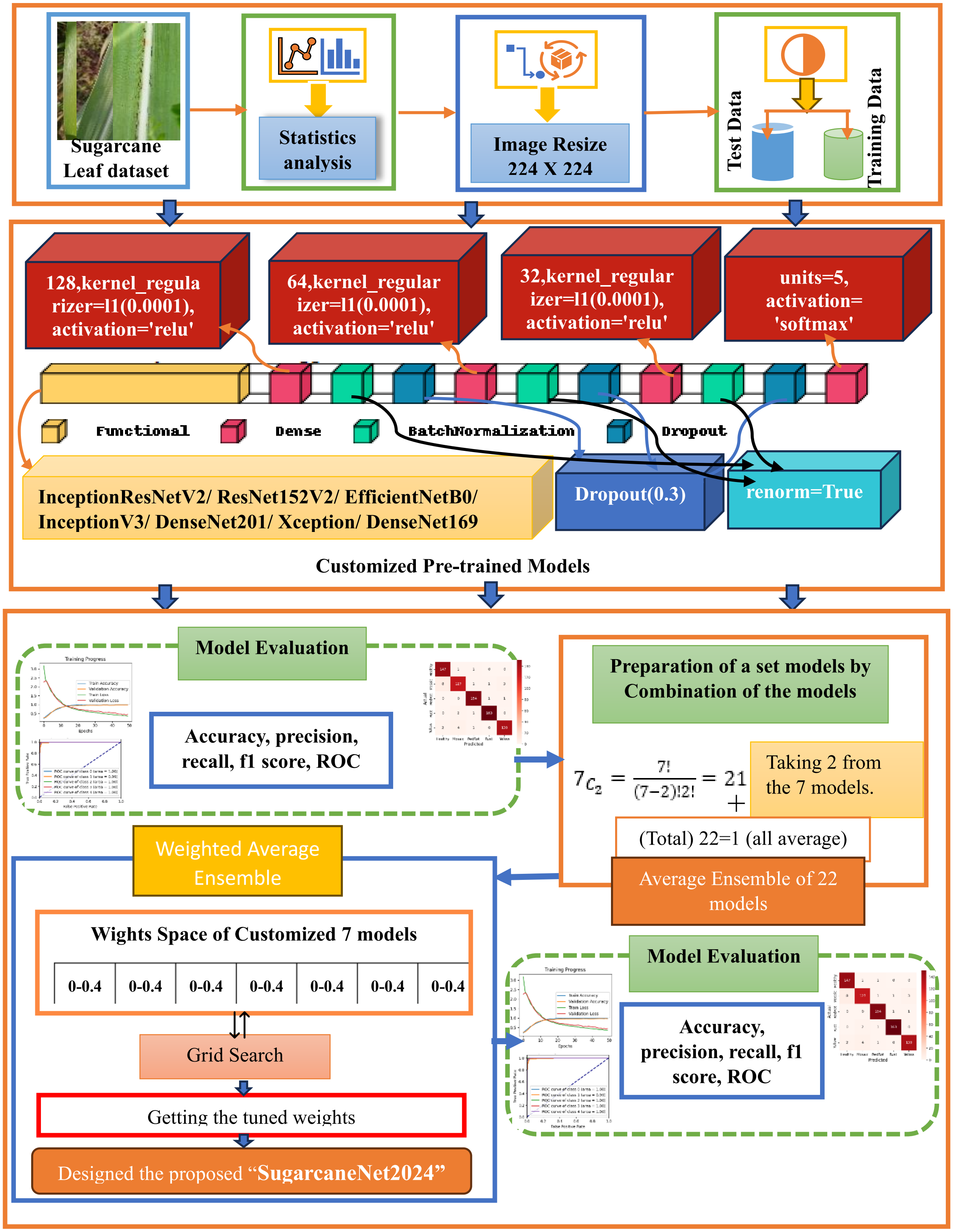

Figure 4. Flow diagram of the proposed "SugarcaneNet" model.

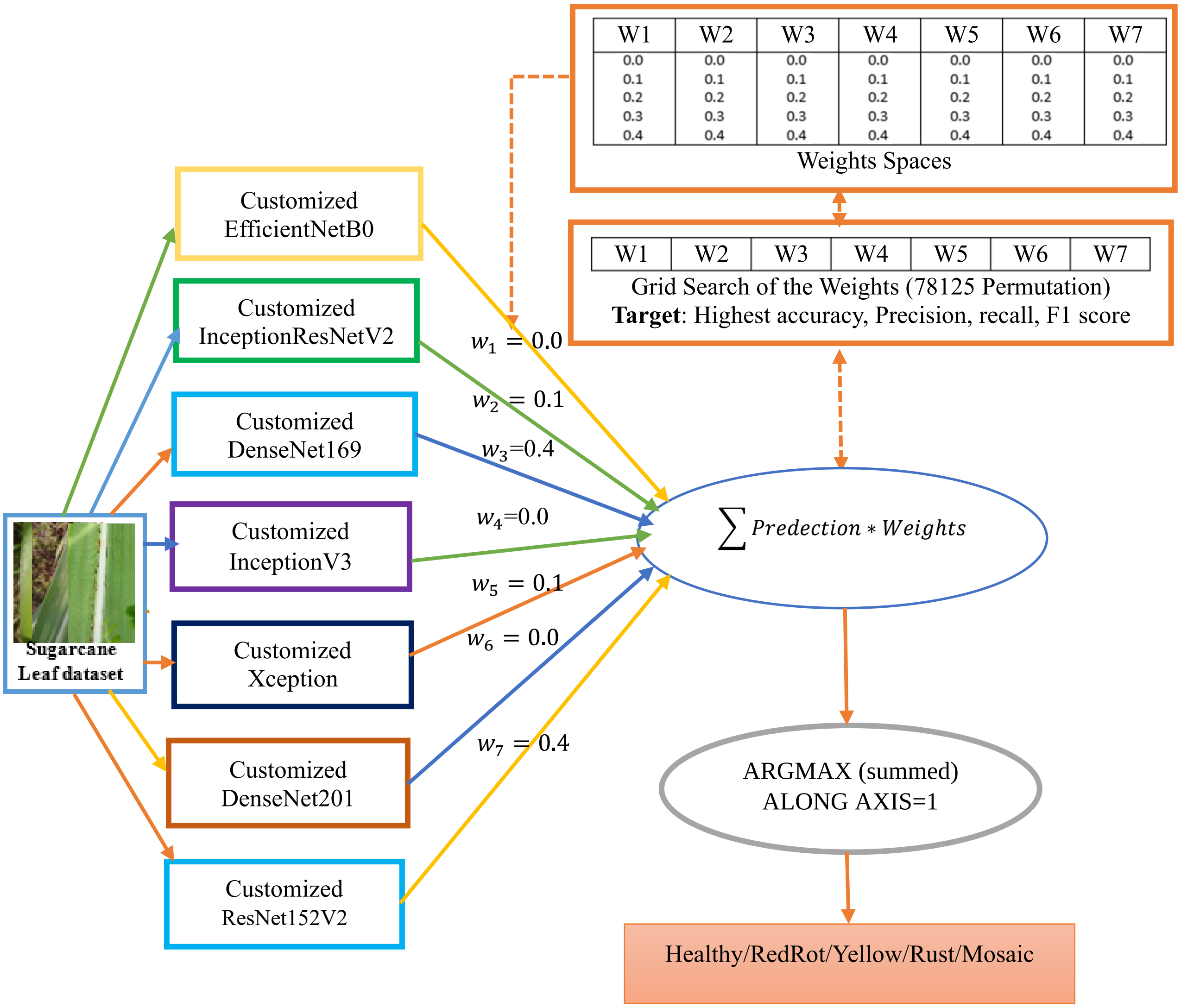

Figure 5. Proposed SugarcaneNet model.

Table 4. Performance metrics.

| | |
|---|---|
| Accuracy | $\frac{TP+TN}{TP+TN+FP+FN}*100$ |
| Precision | $\frac{TP}{TP+FP}*100$ |
| Recall | $\frac{TP}{TP+FN}*100$ |
| F1 score | $2 * \left(\frac{Precision*Recall}{Precision+Recall}\right) * 100$ |

## 4. Result and Discussion

The performance metrics for various epochs for the customized pre-trained models are presented in Figure 6. The precision, recall, mean square error (MSE), accuracy, and loss, for all the models show an increase in precision, recall, and accuracy with increasing epochs up to 50, while MSE and loss show a decrease with increasing epochs towards zero in the same epoch range. Notably, early stop optimization limits the customized InceptionResNetV2 model to only 13 epochs. This model's validation loss has a distinct pattern, first dropping to 1.63 after 7 epochs then rapidly increasing to 9.15 by the 11$^{th}$ epoch. The precision, recall, and MSE values are 0.9547, 0.9484, and 0.0038, respectively, while the validation loss stabilizes at 1.85 at the last epoch. Following the last epoch, the performance metrics for the customized ResNet152V2 model are as follows: The results show that the accuracy, precision, recall, and MSE values were 0.9473, 0.9481, 0.9429, and 0.0182, respectively. The validation loss was 0.4006, the accuracy, precision, recall, and MSE values were 0.1732, 0.9967, 0.9967, and recall. Over the course of 50 epochs, the customized efficientNetB0 model shows remarkably smooth curves. The final training result displays a training loss of 0.3537, training accuracy of 0.9977, training precision of 0.9989, training recall of 0.9960, and training MSE of 0.0011. Accuracy, precision, recall, and MSE values are 0.9643, 0.9668, 0.9630, and 0.0107, respectively, with a validation loss of 0.4418. In a similar vein, the last epoch of the customized InceptionV3 model yields the following results: training loss of 0.2695, training accuracy, precision, recall, and MSE of 0.0021. Accuracy, precision, recall, and MSE values are 0.9657, 0.9668, 0.9630, and 0.0122, respectively, with a validation loss of 0.3844. Additionally, the performance metrics following the last epoch of the customized DenseNet201 model are as follows: with accuracy, precision, recall, and MSE values of 0.9660, 0.9670, 0.9660, and 0.0129, respectively; validation loss: 0.3260; loss: 0.1191, accuracy: 0.9995, precision: 0.9995, recall: 0.9995, and MSE: 2.3563e-04.

Furthermore, the results for the customized Xception model following the final epoch are as follows: validation loss: 0.2774, MSE: 6.5263e-05, accuracy, precision, recall, and loss: 0.1288, with values for accuracy, precision, recall, and MSE of 0.9696, 0.9695, 0.9670, and 0.0111, respectively. Lastly, the following performance metrics are obtained from the customized DenseNet169 model after its last epoch: Precision, accuracy, recall, and MSE values are 0.9511, 0.9523, 0.9498, and 0.0163, respectively. Validation loss is 0.4143, loss is 0.2131, accuracy is 0.9955, precision is 0.9955, recall is 0.9949, and MSE is 0.0017.

However, other models show consistent performance up to 50 epochs with varied final metrics, while InceptionResNetV2 stops at 13 epochs with fluctuating validation loss.

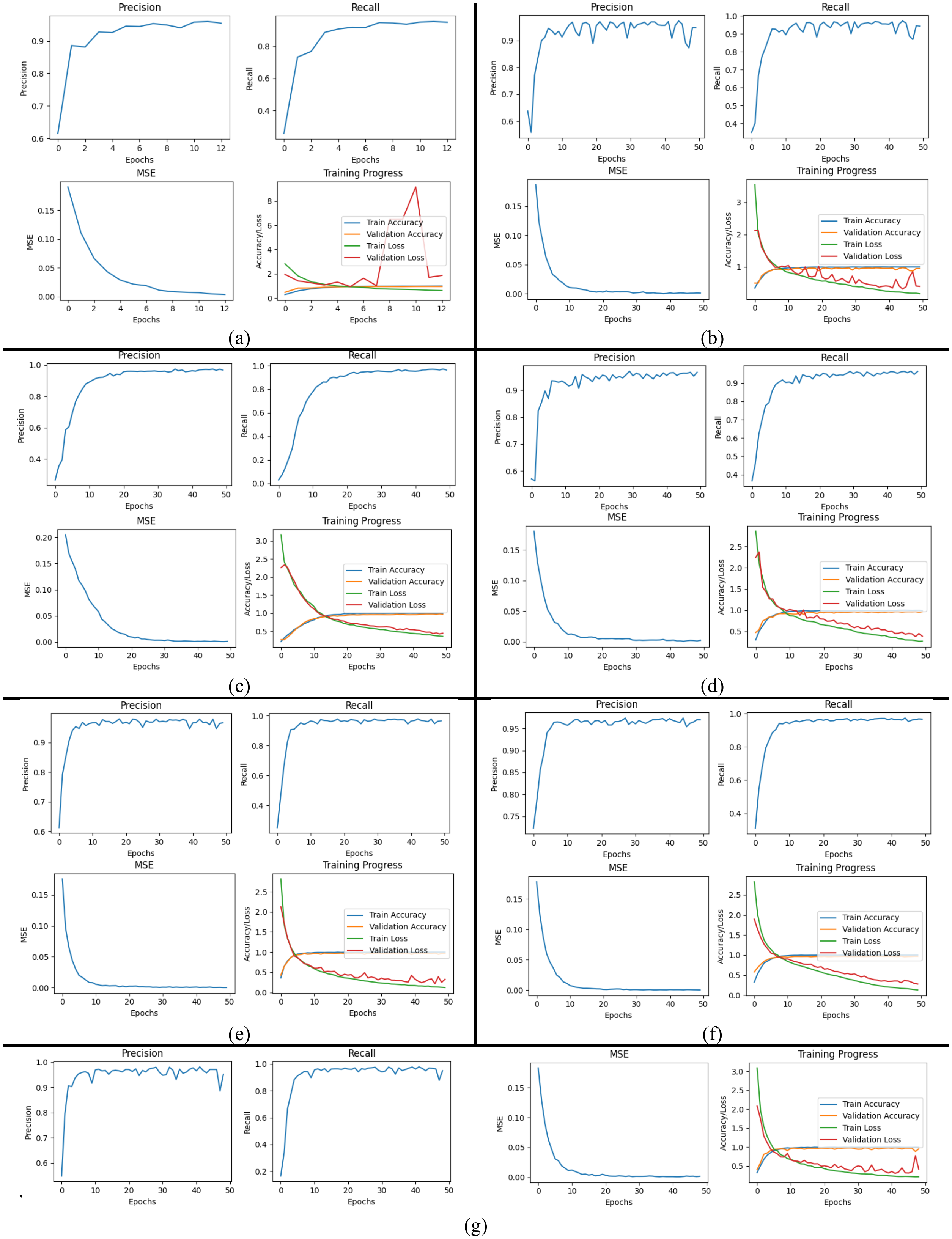

Figure 6. Precision, recall, MSE, accuracy and loss curves of customized models (a) InceptionResNetV2;(b) ResNet152V2; (c) EfficientNetB0; (d) InceptionV3; (e) DenseNet201; (f) Xception; (g) DenseNet169.

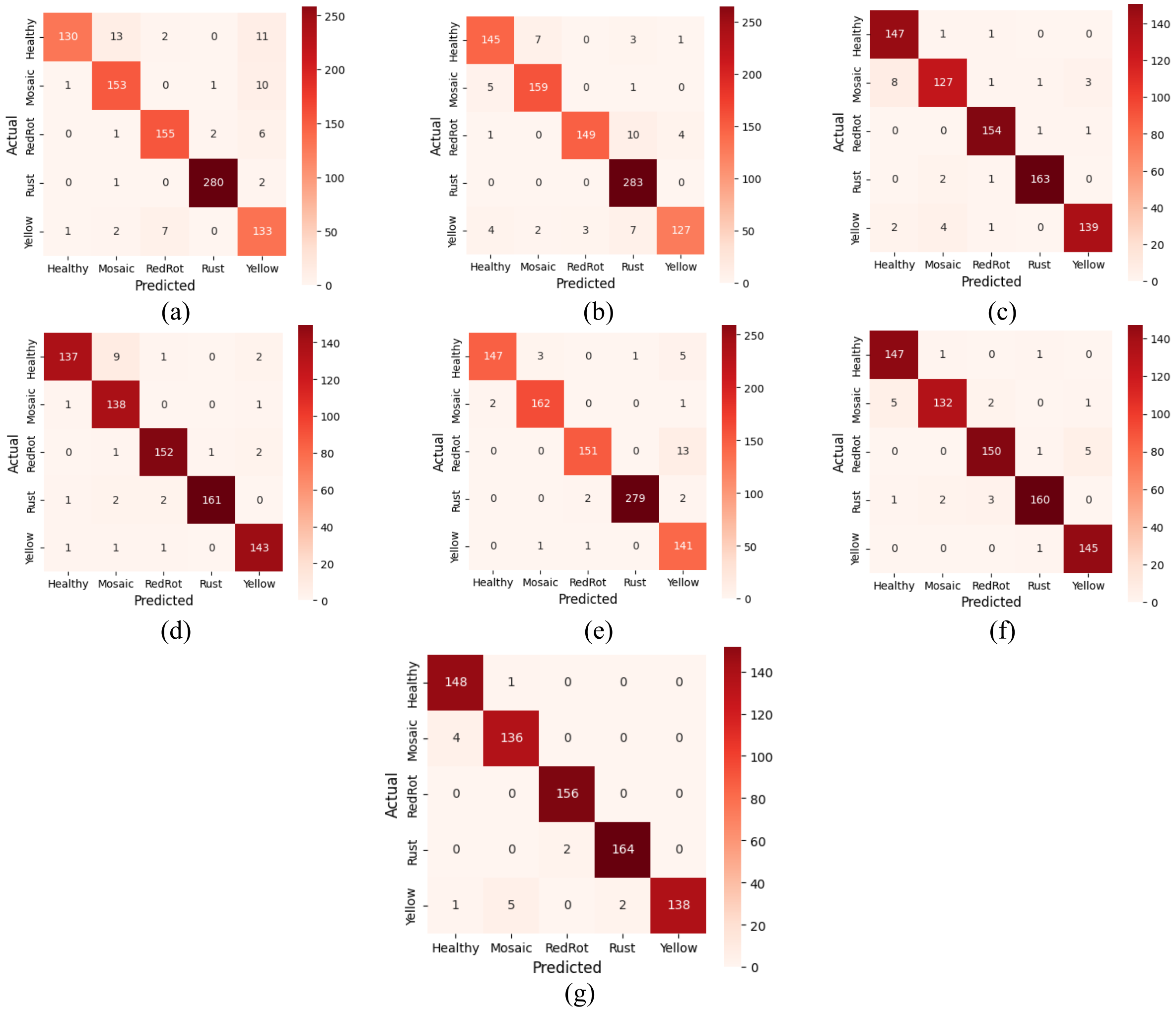

Figure 7. Confusion Matrix of customized model (a) InceptionResNetV2;(b) ResNet152V2; (c) EfficientNetB0; (d) InceptionV3; (e) DenseNet201; (f) Xception; (g) DenseNet169.

The confusion matrix, based on predictions of 757 test data instances, is shown in Figure 7 and describes the performance of improved pre-trained models in sugarcane leaf disease classification. True positives, false positives, true negatives, and false negatives are all represented in the matrix for each model. InceptionResNetV2 had the lowest classification accuracy among the adjusted models that identified 697 out of 757 unknown sugarcane leaf diseases properly. The rest of the models, nevertheless, such the modified ResNet152V2, EfficientNetB0, InceptionV3, DenseNet201, Xception, and DenseNet169, showed steadily better results. Modified DenseNet169 obtained the maximum classification performance that identified 742 out of 757 unknown data items properly. To comprehensively evaluate these models, accuracy, precision, recall, and F1 score were figured out. The results, presented in Tables 5-11, illustrate distinct and relevant performance values, conveying significant insights. The specific accuracy scores for ResNet152V2, InceptionV3, DenseNet201, Xception, DenseNet169, customized InceptionResNetV2, and EfficientNetB0 are 93.41%, 94.73%, 96.43%, 96.56%, 96.59%, 96.96%, and 98.01%, respectively. Significantly better than the others are Modified DenseNet169.

Additionally, the results show that Modified DenseNet169 has the best precision (98), recall (98), and F1 score (98), whereas InceptionResNetV2 has the lowest values (93, 93, and 92). The higher performance of DenseNet169 across all parameters is further highlighted by individual class-wise study.

Table 5. Performance metrics for the InceptionResNetV2.

| Class Names | Precision (%) | Recall (%) | F1-Score (%) | Support | Accuracy (%) |
|---|---|---|---|---|---|
| Healthy | 98 | 83 | 90 | 156 | 93.41 |
| Mosaic | 90 | 93 | 91 | 165 | |
| RedRot | 95 | 95 | 95 | 164 | |
| Rust | 99 | 99 | 99 | 283 | |
| Yellow | 82 | 93 | 87 | 143 | |
| Macro Avg.~93 | | ~93 | ~92 | ~911 | |

Table 6. Performance metrics for ResNet152V2.

| Class Names | Precision (%) | Recall (%) | F1-Score (%) | Support | Accuracy (%) |
|---|---|---|---|---|---|
| Healthy | 94 | 93 | 93 | 156 | 94.73 |
| Mosaic | 95 | 96 | 95 | 165 | |
| RedRot | 98 | 91 | 94 | 164 | |
| Rust | 93 | 100 | 96 | 283 | |
| Yellow | 96 | 89 | 92 | 143 | |
| Macro Avg. ~95 | | ~94 | ~94 | ~911 | |

Table 7. Performance metrics for EfficientNet B0.

| Class Names | Precision (%) | Recall (%) | F1-Score (%) | Support | Accuracy (%) |
|---|---|---|---|---|---|
| Healthy | 94 | 99 | 96 | 149 | 96.43 |
| Mosaic | 95 | 91 | 93 | 140 | |
| RedRot | 97 | 99 | 98 | 156 | |
| Rust | 99 | 98 | 98 | 166 | |
| Yellow | 97 | 95 | 96 | 146 | |
| Macro Avg.~96 | | ~96 | ~96 | ~757 | |

Table 8. Performance metrics for InceptionV3.

| Class Names | Precision (%) | Recall (%) | F1-Score (%) | Support | Accuracy (%) |
|---|---|---|---|---|---|
| Healthy | 98 | 92 | 95 | 149 | 96.56 |
| Mosaic | 91 | 99 | 95 | 140 | |
| RedRot | 97 | 97 | 97 | 156 | |
| Rust | 99 | 97 | 98 | 166 | |
| Yellow | 97 | 98 | 97 | 146 | |
| Macro Avg.~97 | | ~97 | ~97 | ~757 | |

Table 9. Performance metrics for DenseNet201.

| Class Names | Precision (%) | Recall (%) | F1-Score (%) | Support | Accuracy (%) |
|---|---|---|---|---|---|
| Healthy | 99 | 94 | 96 | 156 | 96.59 |
| Mosaic | 98 | 98 | 98 | 165 | |
| RedRot | 98 | 92 | 95 | 164 | |
| Rust | 100 | 99 | 99 | 283 | |
| Yellow | 87 | 99 | 92 | 143 | |
| Macro Avg.~96 | | ~96 | ~96 | ~911 | |

Table 10. Performance metrics for the Xception.

| Class Names | Precision (%) | Recall (%) | F1-Score (%) | Support | Accuracy (%) |
|---|---|---|---|---|---|
| Healthy | 96 | 99 | 97 | 149 | 96.96 |
| Mosaic | 98 | 94 | 96 | 140 | |
| RedRot | 97 | 96 | 96 | 156 | |
| Rust | 98 | 96 | 97 | 166 | |
| Yellow | 96 | 99 | 98 | 146 | |
| Macro Avg.~97 | | ~97 | ~97 | ~757 | |

Table 11. Performance metrics for DenseNet169.

| Class Names | Precision (%) | Recall (%) | F1-Score (%) | Support | Accuracy (%) |
|---|---|---|---|---|---|
| Healthy | 97 | 99 | 98 | 149 | 98.01 |
| Mosaic | 96 | 97 | 96 | 140 | |
| RedRot | 99 | 100 | 99 | 156 | |
| Rust | 99 | 99 | 99 | 166 | |
| Yellow | 100 | 95 | 97 | 154 | |
| Macro Avg~98 | | ~98 | ~98 | ~757 | |

Figure 8 shows the Receiver Operating Characteristic (ROC) curve of all the customized pre-trained models. Figure 8 shows how the true positive rate (TPR) and false positive rate (FPR) relate to one another at various threshold values. Simplified as the ratio of true positives to total positives (true positives plus false negatives), TPR, or sensitivity, is a measure of this ratio. The number of false positives divided by the total number of real negatives is how the false positive rate (FPR) is determined, on the other hand. Understanding how well the model can distinguish between positive and negative examples in all categories can be gained from looking at the micro-average ROC curve. Analyzing the curve's shape and proximity to the top-left corner, as well as the area under the curve (AUC), which has a value of 1 denoting perfect discrimination, enables a comprehensive evaluation of the model's efficacy. All pre-trained models have consistently produced an AUC near 1 in every class, except for InceptionResNetV2. This means all models are capable of correctly classifying all forms of sugarcane leaf diseases.

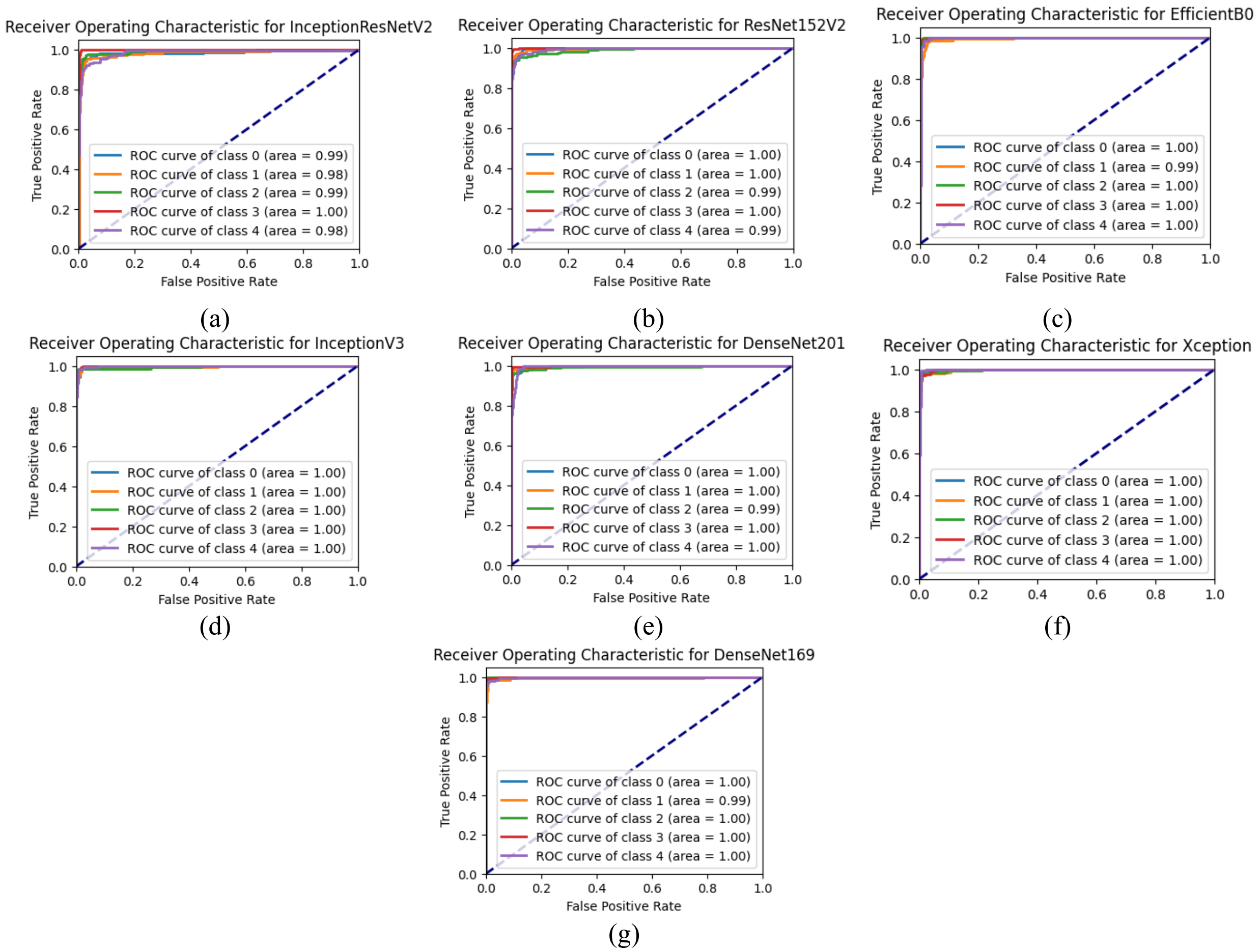

Figure 8. ROC of customised models (a) InceptionResNetV2;(b) ResNet152V2; (c) EfficientNetB0; (d) InceptionV3; (e) DenseNet201; (f) Xception; (g) DenseNet169.

The average ensemble strategy is the next step. We are able to create several ensemble models by merging two, three, four, five, six, and seven of our seven customized pre-trained models.
That means total can be calculated as follows as

$$7_{C_2} + 7_{C_3} + 7_{C_4} + 7_{C_5} + 7_{C_6} + 7_7 = \frac{7!}{(7-2)!2!} + \frac{7!}{(7-3)!3!} + \frac{7!}{(7-4)!4!} + \frac{7!}{(7-5)!5!} + \frac{7!}{(7-6)!6!} + \frac{7!}{(7-7)!7!} = 120$$

However, because of its simplicity, we only selected the ensemble model, which combines the two and all (seven) models. This results in an average ensemble model of 21+1=22. Table 12 presents a list of the models' outcomes. The accuracy, precision, recall, and F1 score of the two ensemble models-InspectionResNetV2 + DenseNet201 and InspectionResNetV2 + ResNet152V2-are all lower than those of the prior customized DenseNet169 model. However, compared to the prior pre-trained model, the rest of the 20 average ensemble model exhibits better accuracy.
DenseNet169 + ResNet152V2 and EfficientNetB0 + InceptionV3 are the two mixed ensemble models that perform the best. The results show accuracy, recall, and precision of 99% and an F1 score of 99.23%. The ensemble of all models (EfficientNetB0+ InceptionResNetV2+ DenseNet169+ InceptionV3+Xecption + DenseNet201+ResNet152V2) achieved an impressive result with an F1 score of 99.45%, 100% precision, and 99% recall.

To further improve performance, we carried out the optimized weighted average ensemble of all models. The best optimized weights gotten from grid search are listed in Table 13. Using the weights, our proposed sugarcanNet2024 has achieved the most significant performance. It provides the highest accuracy over all kinds of experiments and previous studies [45]. This provides 99.67% accuracy, 100% precision, recall, and an F1 score.

There are a lot of models when using the ensemble method. For this reason, we have excluded the confusion matrix, ROC curves, and test data visualization from this section. Figures 9, 10, and 11 only display our most well-performed proposed model (SugarcaneNet) in terms of the confusion matrix, ROC curve, and results. The SugarcaneNet model is able to classify 754 out of 757 images of sugarcane leaf diseases. There are just three that are incorrectly categorized. For every class, the Roc curve displays auc 1. As can be seen from the visual result in Figure 11, the model classifies the test data with high confidence. After examining every conceivable parameter, our suggested SugarcaneNet model is the most effective in identifying and categorizing sugarcane leaf diseases.

12. Performance analysis of ensemble models and proposed models.

| **Types of Ensembles** | **Average Ensemble Model's Name** | **Precision (%)** | **Recall (%)** | **F1 Score (%)** | **Accuracy (%)** |
|---|---|---|---|---|---|
| Average Ensemble | EfficientNetB0 + InceptionResnetV2 | 99 | 99 | 99 | 98.68 |
| | EfficientNetB0 + DenseNet169 | 99 | 99 | 99 | 99.00 |
| | EfficientNetB0 + InceptionV3 | 99 | 99 | 99 | 99.23 |
| | EfficientNetB0 + Xception | 99 | 99 | 99 | 99.00 |
| | EfficientNetB0 + DenseNet201 | 99 | 99 | 99 | 98.90 |
| | EfficientNetB0 + ResNet152V2 | 99 | 99 | 99 | 98.79 |
| | InspectionResNetV2 + DeseNet169 | 98 | 99 | 99 | 98.68 |
| | InspectionResNetV2 + InceptionV3 | 98 | 98 | 98 | 98.24 |
| | InspectionResNetV2 + Xception | 98 | 98 | 98 | 98.24 |
| | InspectionResNetV2 + DenseNet201 | 98 | 98 | 98 | 97.80 |
| | InspectionResNetV2 + ResNet152V2 | 96 | 96 | 96 | 96.04 |
| | Xception + DenseNet201 | 99 | 99 | 99 | 99.00 |
| | Xception + ResNet152V2 | 99 | 99 | 99 | 98.90 |
| | DenseNet201+ ResNet152V2 | 98 | 98 | 98 | 98.13 |
| | DenseNet169+ InceptionV3 | 99 | 99 | 99 | 98.79 |
| | DenseNet169+ Xception | 99 | 99 | 99 | 98.90 |
| | DenseNet169+ DenseNet201 | 99 | 99 | 99 | 98.90 |
| | DenseNet169+ ResNet152V2 | 99 | 99 | 99 | 99.23 |
| | InceptionV3+Xception | 99 | 99 | 99 | 98.90 |
| | InceptionV3+Densenet201 | 99 | 98 | 99 | 98.57 |
| | InceptionV3+ResNet152V2 | 99 | 99 | 99 | 98.79 |
| | EfficientNetB0+ InceptionResNetV2+ DenseNet169+ InceptionV3+Xecption + DenseNet201+ResNet152V2 | 100 | 99 | 99 | 99.45 |
| Optimized Weighted Average Ensemble | *Proposed Model (SugarcaneNet) | **100** | **100** | **100** | **99.67** |

*Weighted average ensemble of EfficientNetB0, InceptionResNetV2, DenseNet169, InceptionV3, Xecption, DenseNet201, ResNet152V2.

Table 13. Tuned weights values.

| Model Name | EfficientNetB0 | InceptionResNetV2 | DenseNet169 | InceptionV3 | Xecption | DenseNet201 | ResNet152V2 |
|---|---|---|---|---|---|---|---|
| Weight Values | 0.0 | 0.1 | 0.4 | 0.0 | 0.1 | 0.0 | 0.4 |

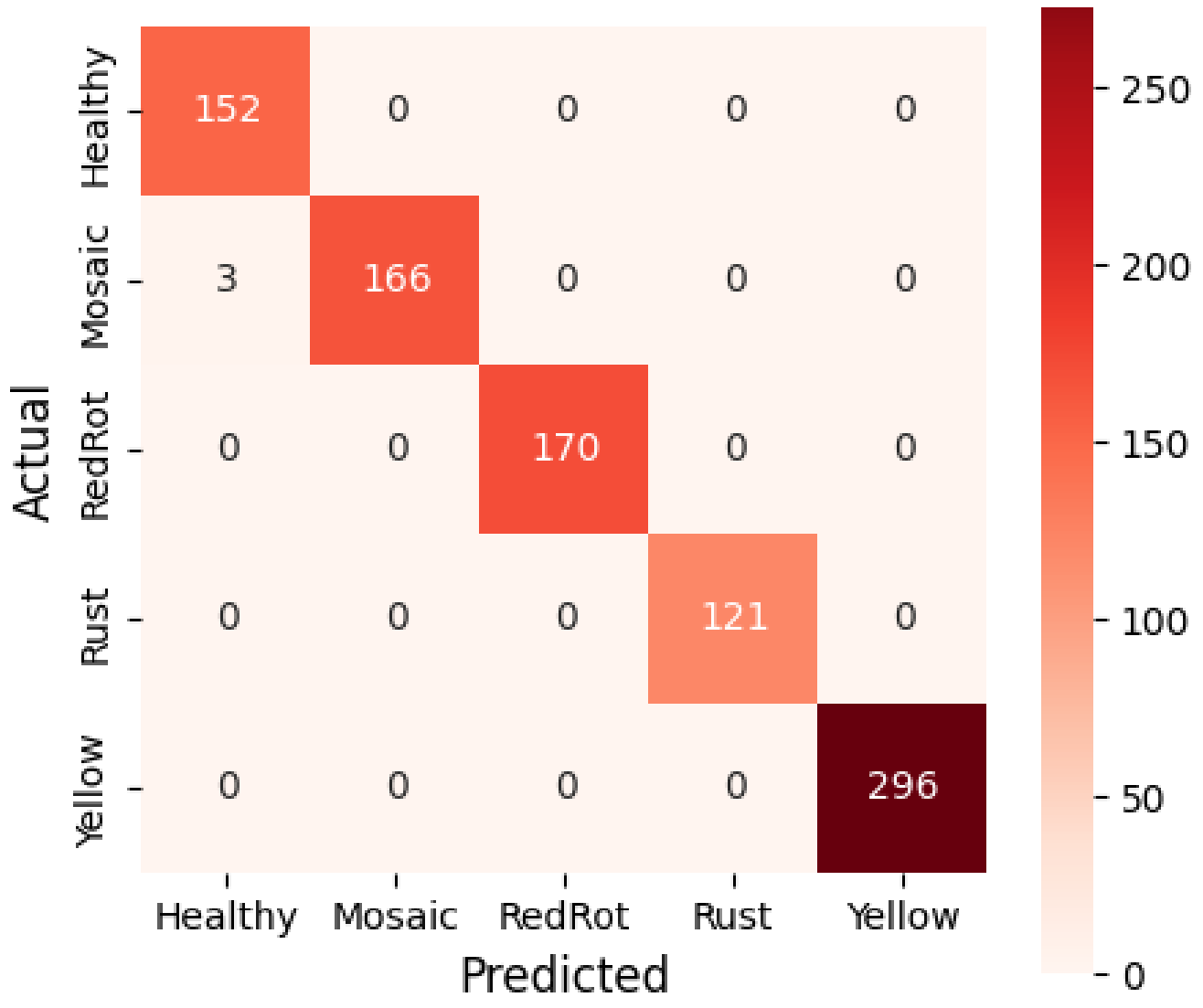

Figure 9. Confusion Matrix of “SugarcaneNet”.

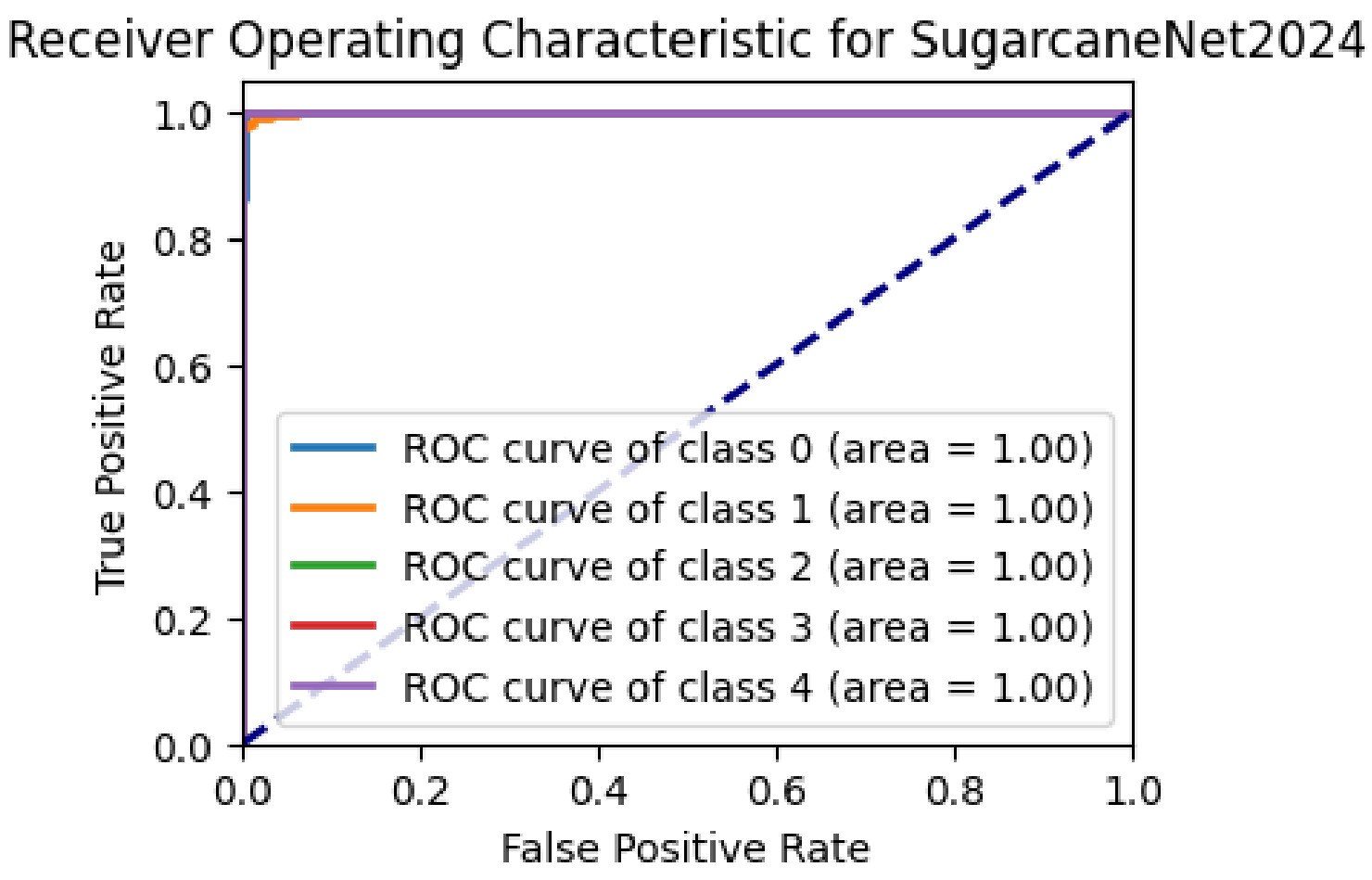

Figure 10. ROC curve of “SugarcaneNet”.

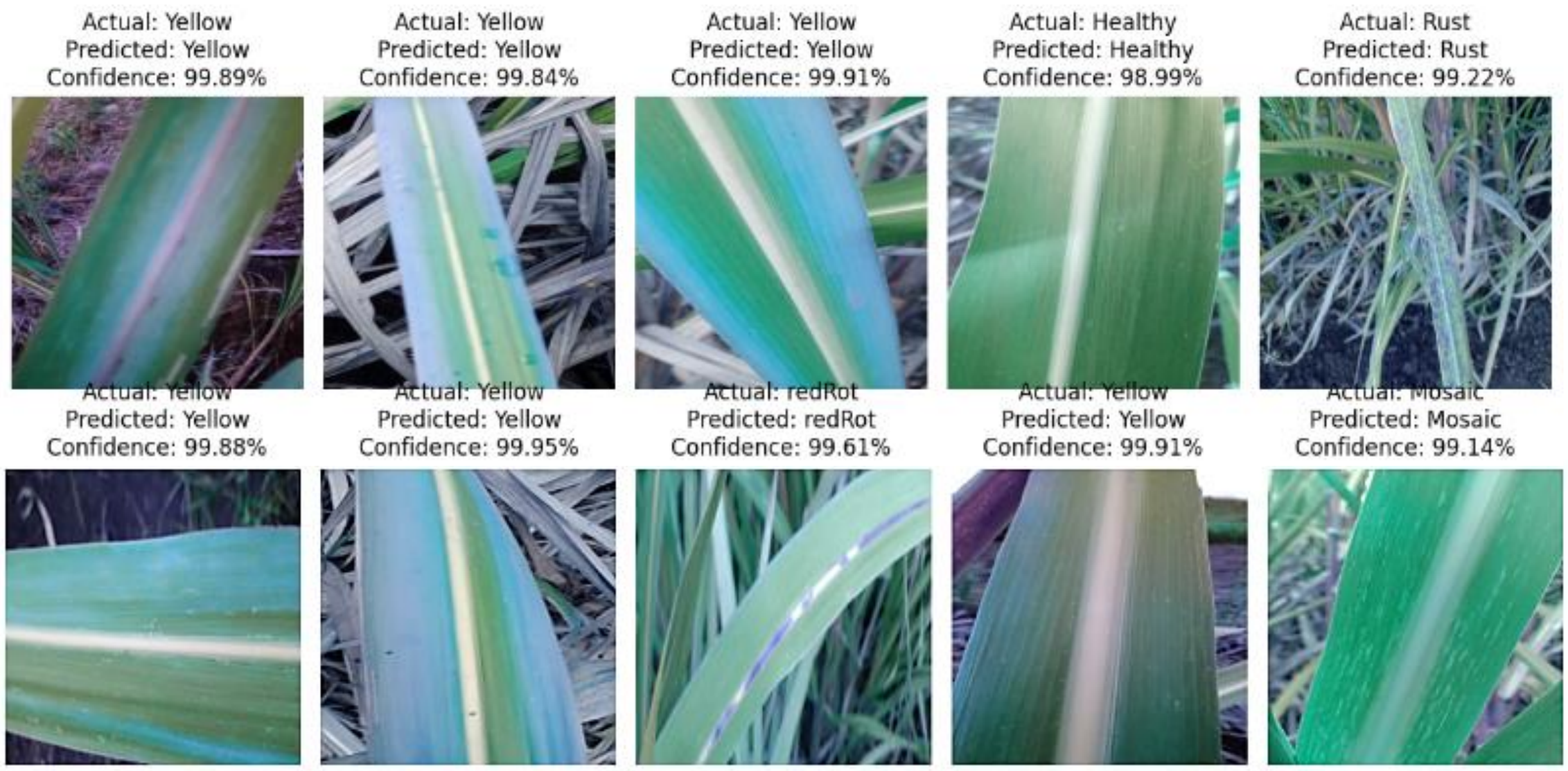

Figure 11. Final predicted output with confidence.

## 5. Conclusion and Future work

Our study tackles a problem that the worldwide sugar industry is currently facing: the prompt identification and treatment of diseases that affect sugarcane. To provide farmers and agricultural specialists with a vital tool, our research concentrated on creating SugarcaneNet, an advanced model that automates the diagnosis of various diseases. The study performs a thorough comparative analysis, assessing the performance of seven modified, regularized pre-trained models and 22 average ensemble models that are derived from themselves. Risks associated with overfitting are reduced by using strategies like LASSO, Dropout, and "renorm" regularization. Individually customized models perform significantly better than their predecessors, with the average ensemble showing a notable advance.

In the end, sugarcaneNet, a fine-tuned weighted average ensemble, is developed and implemented for future improvement. This ensemble greatly improves performance measures with outstanding accuracy, precision, recall, and F1 scores of 99.67%, 100%, 100%, and 100%, respectively.

Although our study concentrated on the identification of five different sugarcane disease categories, there is still plenty of room for future research to build on this framework. Through the incorporation of a wider range of sugarcane diseases and the implementation of our SugarcaneNet model in actual agricultural environments, scientists may enhance and verify its efficacy. Moreover, putting sugarcaneNet to use in field tests and agricultural activities would offer important insights into its usefulness and efficacy in actual settings. Researchers may assess the effectiveness of the model and direct future improvements by collaborating with farmers and agricultural organizations to collect data on disease prevalence, treatment outcomes, and overall crop health.

However, establishing sugarcaneNet is a big step in ensuring the productivity and sustainability of sugarcane farming. The timely identification of diseases allows for the implementation of mitigation and intervention plans, which in turn reduces yield losses and maintains sugarcane farming's economic sustainability.

**Author Contribution:** MSHT: Implementation, Formal Analysis, and original drafting. SA: Writing Methodology and data collection. AHN: Writing Introduction and Literature Survey. MA: Formatting, Validation and funding, RBS: Reviewing and Editing; AFA: Writing and mentoring

**Declaration of Competing Interest:** The authors declare no conflict of interest.

**Funding:** This research received no external funding.

**Data Availability**: Publicly available. Link: https://data.mendeley.com/datasets/9424skmnrk/1